%% file: main.tex
\documentclass[10pt,twocolumn,letterpaper]{article}

\usepackage[pagenumbers]{cvpr} 

\usepackage{float}
\usepackage{multirow}
\usepackage[accsupp]{axessibility} 
\input{preamble}

\definecolor{cvprblue}{rgb}{0.21,0.49,0.74}
\usepackage[pagebackref,breaklinks,colorlinks,allcolors=cvprblue]{hyperref}
\usepackage[most]{tcolorbox}
\usepackage{booktabs}
\usepackage{tabularx}
\usepackage{threeparttable}

\usepackage{times}
\usepackage[T1]{fontenc}

\def\paperID{17} 
\def\confName{CVPR}
\def\confYear{2026}

\title{ARM: Advantage Reward Modeling for Long-Horizon Manipulation}

\author{
    Yiming Mao$^1$ \qquad 
    Zixi Yu$^{1, 2}$ \qquad 
    Weixin Mao$^{1, \dagger}$ \qquad 
    Yinhao Li$^1$ \\
    Qirui Hu$^1$ \qquad 
    Zihan Lan$^1$ \qquad 
    Minzhao Zhu$^1$ \qquad 
    Hua Chen$^{1, 3, \ast}$ \\
    \vspace{3pt}
    {\small $^1$LimX Dynamics \qquad $^2$Beijing University of Posts and Telecommunications \qquad $^3$Zhejiang University} \\
    {\tt\small \{aiming, nemo, waynemao, mason, ryan.hu, sober, mayer\}@limxdynamics.com} \\
    {\tt\small huachen@intl.zju.edu.cn}
    \thanks{$^\ast$Corresponding author. $^\dagger$Project leader.}
}

\begin{document}
\maketitle
\input{sec/0_abstract}    
\input{sec/1_intro}
\input{sec/2_rela_work}

\input{sec/3_method}

\input{sec/4_exp}
\input{sec/5_conclusion}
\clearpage
{
    \small
    \bibliographystyle{ieeenat_fullname}
    \bibliography{main}
}

\input{sec/X_suppl}

\end{document}

%% file: preamble.tex
\usepackage[dvipsnames,table]{xcolor} 
\newcommand{\progressive}{\textcolor{ForestGreen}{\textbf{Progressive}}}
\newcommand{\regressive}{\textcolor{pink}{\textbf{Regressive}}}
\newcommand{\stagnant}{\textbf{Stagnant}}

\usepackage{array}



%% file: sec/0_abstract.tex
\begin{abstract}
Long-horizon robotic manipulation remains challenging for reinforcement learning (RL) because sparse rewards provide limited guidance for credit assignment. Practical policy improvement thus relies on richer intermediate supervision, such as dense progress rewards, which are costly to obtain and ill-suited to non-monotonic behaviors such as backtracking and recovery.
To address this, we propose Advantage Reward Modeling (ARM), a framework that shifts from hard-to-quantify absolute progress to estimating relative advantage. We introduce a cost-effective tri-state labeling strategy---\textit{Progressive}, \textit{Regressive}, and \textit{Stagnant}---that reduces human cognitive overhead while ensuring high cross-annotator consistency. By training on these intuitive signals, ARM enables automated progress annotation for both complete demonstrations and fragmented DAgger-style data. Integrating ARM into an offline RL pipeline allows for adaptive action-reward reweighting, effectively filtering suboptimal samples. Our approach achieves a 99.4\% success rate on a challenging long-horizon towel-folding task, demonstrating improved stability and data efficiency over current VLA baselines with near-zero human intervention during policy training.

\vspace{1em} 
\noindent \textbf{Project page:} \url{https://aiming1998.github.io/ARM}
\end{abstract}

%% file: sec/1_intro.tex
\section{Introduction}
\label{sec:intro}
\begin{figure*}[htbp]
    \centering
    \includegraphics[width=0.95\linewidth]{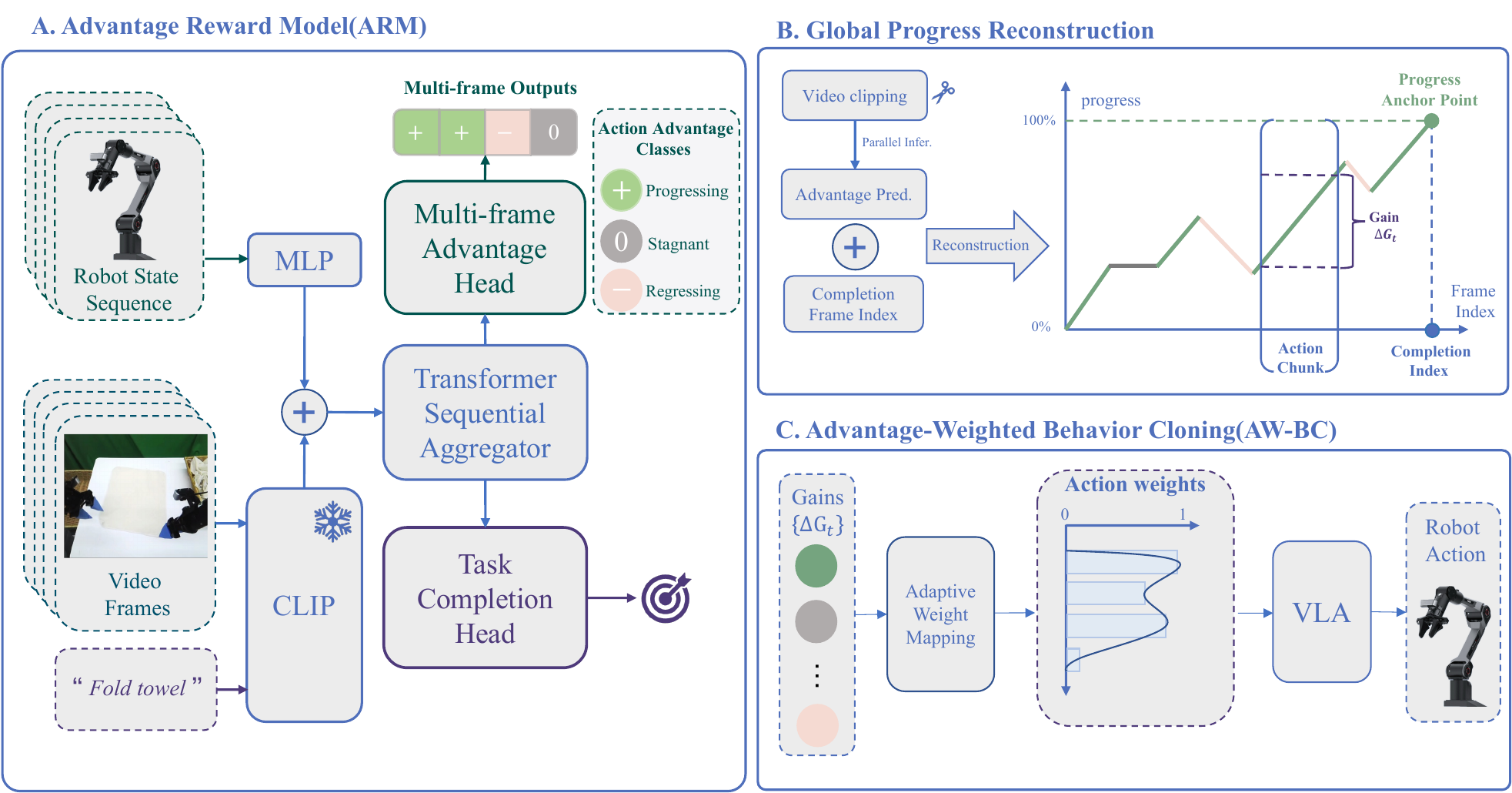}
    \caption{\textbf{Overview of our proposed framework.} The system consists of three main components: (1) The Advantage Reward Model (ARM) with its MIMO-based Temporal Transformer, supervised by a lightweight tri-state labeling strategy; (2) An automated pipeline for global progress reconstruction; and (3) The Advantage-Weighted Behavior Cloning (AW-BC) algorithm, which optimizes the policy using length-invariant relative gains extracted from the reconstructed progress.}
    \label{fig:arm_architecture}
\end{figure*}

The rapid evolution of Vision-Language-Action (VLA) models~\cite{openvla, gr00tn1, black2024pi_0, pi05} has advanced general-purpose robotic manipulation.
However, most existing VLA approaches rely heavily on imitation learning (IL)~\cite{Osa_2018}, which demands massive datasets and incurs considerable human labor and physical resources costs during large-scale data collection~\cite{o2024open, khazatsky2024droid, bu2025agibot, walke2024bridgedatav2datasetrobot, wu2025robomind, hou2025robomind2}.
Beyond data quantity, the inherent suboptimality and noise in human demonstrations, especially in complex, long-horizon tasks, often impede policy convergence. Reinforcement Learning (RL)~\cite{rl} provides a promising alternative by enabling autonomous policy refinement beyond expert demonstrations~\cite{pi06,li2025gr}.

Nevertheless, effective RL in long-horizon manipulation hinges on informative reward signals. While sparse rewards (e.g., binary success indicators) are straightforward to specify, they struggle to yield effective learning signals, frequently leading to convergence difficulties in long-horizon manipulation tasks. Consequently, high-quality dense rewards or informative value functions are essential to provide continuous supervision and facilitate effective credit assignment.

Current frameworks~\cite{pi06, chen2025sarmstageawarerewardmodeling} attempt to leverage dense signals through advantage estimation or sample reweighting. However, they depend on high-precision progress reward models to mitigate the notorious credit assignment problem. This dependency constitutes a pervasive ``Reward Engineering Bottleneck'', limiting both scalability and stability of VLA deployment in unstructured environments. Designing a cost-effective reward function that provides stable and high-frequency feedback remains a formidable challenge. In particular, existing evaluation paradigms predicated on absolute progress are limited by several critical bottlenecks~\cite{pi06, gvl, wu2026largerewardmodelsgeneralizable, liang2026robometer, tan2025robodopaminegeneralprocessreward, zhang2025rewindlanguageguidedrewardsteach, chen2025sarmstageawarerewardmodeling}:
First, Zero-shot VLMs suffer from considerable unreliability and prohibitive costs; they not only incur high inference overhead, but also yield low-precision annotations due to their lack of spatial-geometric grounding, which manifests as non-monotonic oscillations in reward signals~\cite{gvl,sontakke2023roboclip,chen2025sarmstageawarerewardmodeling}. 
Second, current schemes exhibit quantization ambiguity in failure states. By predicating progress modeling on a strict monotonicity assumption and relying on simplistic video rewinding~\cite{zhang2025rewindlanguageguidedrewardsteach,chen2025sarmstageawarerewardmodeling,tan2025robodopaminegeneralprocessreward} to simulate regression, these methods fail to comprehensively characterize authentic, non-linear operational errors.
Moreover, the conventional reliance on coarse subtask partitions~\cite{chen2025sarmstageawarerewardmodeling, tan2025robodopaminegeneralprocessreward} fails to capture the subtle intra-stage transitions essential for long-horizon tasks—such as critical recovery and corrective maneuvers~\cite{hu2025rac}—ultimately yielding misaligned reward signals and erratic policy updates.

To address these challenges, we introduce Advantage Reward Modeling (ARM).
Our core insight is that defining absolute progress necessitates \textit{ad-hoc}, task-specific heuristics that are difficult to scale. 
In contrast, the relative advantage between states provides a more intuitive, concise, and task-agnostic primitive for annotation. While the recent work VLAC~\cite{zhai2025vlac} also employs interval gain prediction, its methodology is predicated on the assumption of a positive correlation between task progress and time. By decoupling progress rewards from global temporal anchors, ARM naturally accommodates regressive behaviors and error recovery. 
Our core contributions are as follows:
\begin{itemize}
    \item \textbf{Tri-state Advantage Labeling Strategy:} We introduce a labeling method based on three fundamental categories: \progressive, \regressive, and \stagnant. This scheme is task-agnostic, imposes low cognitive load, and is natively compatible with heterogeneous and fragmented datasets.
    \item \textbf{Advantage Reward Model (ARM):} We develop a multimodal reward model that integrates temporal video sequences with robotic proprioceptive states to estimate the relative progress gain of trajectory segments. By anchoring these predictions with a task-completion head, ARM can automatically reconstruct globally consistent dense progress trajectories from discrete tri-state labels.
    \item \textbf{Advantage-Weighted Behavior Cloning (AW-BC):} 
    We extend the Reward-Aligned Behavior Cloning (RA-BC) paradigm~\cite{chen2025sarmstageawarerewardmodeling} by incorporating adaptive scaling coefficients to ensure compatibility with fragmented DAgger data~\cite{dagger}. By leveraging predicted interval gains for advantage-aware reweighting, AW-BC effectively filters suboptimal samples and prioritizes high-value recovery trajectories. Empirically, our framework achieves a near-perfect success rate of 99.4\% on the challenging, long-horizon \textit{towel-folding} task, marking a notable advancement in VLA policy refinement.
\end{itemize}

%% file: sec/2_rela_work.tex
\section{Related Work}

\subsection{Reward for Manipulation}

Traditional reinforcement learning (RL) relies heavily on manual reward shaping, which is often labor-intensive and task-specific.
To mitigate this, inverse reinforcement learning (IRL)~\cite{ng2000algorithms} and learning from human feedback (RLHF)~\cite{NIPS2017_d5e2c0ad} infer reward functions but suffer from identifiability and scalability issues, respectively. 

Vision-language models (VLMs) such as VIP~\cite{ma2022vip} and LIV~\cite{ma2023liv} provide self-supervised goal-distance signals but lack the precision required for fine-grained, contact-rich manipulation. As noted in SARM~\cite{chen2025sarmstageawarerewardmodeling}, single-objective distance metrics fail to capture intermediate progress in long-horizon tasks. A common limitation shared by methods such as GVL~\cite{gvl}, ReWiND~\cite{zhang2025rewindlanguageguidedrewardsteach}, SARM~\cite{chen2025sarmstageawarerewardmodeling}, and VIP~\cite{ma2022vip} is their reliance on a strict monotonicity assumption, which equates task progress with chronological order. However, real-world offline demonstrations often contain mistakes, retries, and temporary regressions, leading to reward misspecification under temporal heuristics. Alternative approaches also present trade-offs: hop-based mechanisms such as Robo-Dopamine~\cite{tan2025robodopaminegeneralprocessreward} sacrifice fine-grained action detail, while zero-shot VLM prompting~\cite{chen2026toprewardtokenprobabilitieshidden,lee2026roborewardgeneralpurposevisionlanguagereward,chen2025elemental} suffers from prediction noise, high latency, and inference cost. To address these issues, we introduce the Advantage Reward Model (ARM), which relaxes temporal monotonicity by evaluating relative progress against historical visual and proprioceptive states, enabling effective advantage estimation even under temporary trajectory deviations.

\subsection{Reward-Aligned Behavior Cloning (RA-BC)}

Learning from suboptimal demonstrations is a critical bottleneck for large-scale robot learning. To address this, the paradigm of Reweighted Behavior Cloning (BC) has been widely explored. Originating from classic baselines like Advantage-Weighted Regression (AWR)~\cite{peng2019advantageweightedregressionsimplescalable}, Advantage-Weighted Actor-Critic (AWAC)~\cite{nair2021awacacceleratingonlinereinforcement}, and Implicit Q-Learning (IQL)~\cite{kostrikov2021offlinereinforcementlearningimplicit}—these approaches extract improved policies by applying advantage-based scalar weights to suppress suboptimal trajectories. 

However, traditional weighting paradigms are limited by a critical bottleneck: they inherently rely on explicit environment rewards to fit global value functions, which are notoriously inaccessible in vision-based, real-world settings. To bypass this, recent methods like SARM~\cite{chen2025sarmstageawarerewardmodeling} introduced the Reward-Aligned Behavior Cloning (RA-BC) framework, leveraging a stage-aware reward model instead of environment rewards. While effective at mitigating data quality issues, SARM trades the reward bottleneck for a new constraint: it heavily relies on prohibitive manual language annotations. In contrast, our proposed ARM eliminates the need for explicit reward engineering by extracting advantage signals purely through relative progress comparisons.

%% file: sec/3_method.tex
\section{Method}

\subsection{Overview of ARM}
\label{sec:overview}

As illustrated in Fig.~\ref{fig:arm_architecture}, the proposed framework shifts the paradigm from absolute progress modeling to relative advantage estimation. The system comprises three synergistic components:

(A) \textbf{Advantage Reward Model:} A Multi-Input Multi-Output (MIMO) Temporal Transformer designed to capture fine-grained relative advantages from multimodal observations (Fig.~\ref{fig:arm_architecture}A). The model is supervised by a lightweight tri-state labeling scheme that categorizes state transitions into \textit{progressive}, \textit{regressive}, or \textit{stagnant} states, providing a cost-effective and task-agnostic training signal.

(B) \textbf{Global Progress Reconstruction:} An automated pipeline that synthesizes the discrete interval gains predicted by ARM into coherent, globally consistent reward trajectories (Fig.~\ref{fig:arm_architecture}B). This process effectively transforms local relative predictions into dense, high-fidelity progress signals suitable for downstream learning.

(C) \textbf{Policy Optimization via AW-BC:} The AW-BC framework that integrates the reconstructed rewards for discriminative sample reweighting (Fig.~\ref{fig:arm_architecture}C). By leveraging length-adaptive gains to prioritize high-value recovery behaviors and filter suboptimal segments, AW-BC facilitates stable offline RL-style policy refinement on noisy, heterogeneous datasets.

\vspace{0.5em}
\subsection{Advantage Reward Modeling}
\label{sec:arm_modeling}
\begin{figure}[t]
  \centering
  \includegraphics[width=\linewidth]{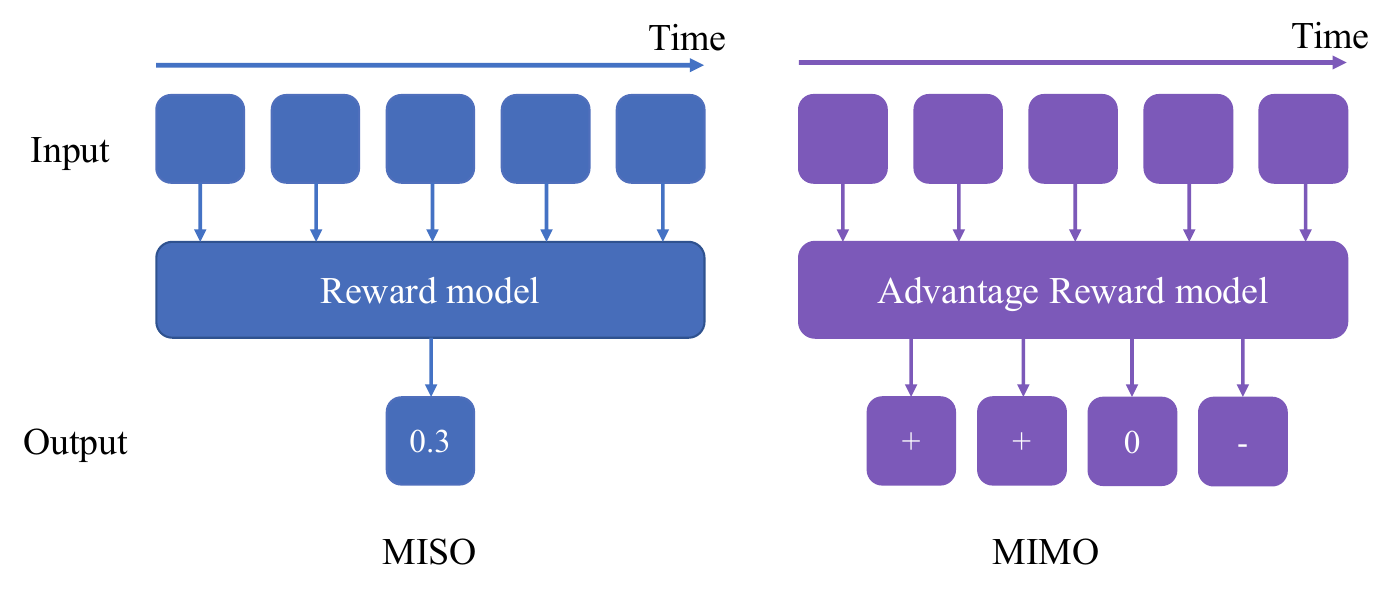}
  \caption{\textbf{Comparison between MISO and MIMO architectures.} MISO stands for Multi-Input Single-Output, and MIMO stands for Multi-Input Multi-Output.}
  \label{miso_mimo}
\end{figure}

The Advantage Reward Model (ARM) is designed to resolve the perceptual ambiguities inherent in isolated frames by shifting the reward estimation paradigm from absolute progress regression to relative advantage classification. Unlike traditional ``Multi-Input Single-Output'' (MISO) models~\cite{chen2025sarmstageawarerewardmodeling, gvl} that collapse temporal context into a single scalar, ARM formulates reward estimation as a Multi-Input Multi-Output (MIMO) sequence learning problem (Fig.~\ref{miso_mimo}). This design allows the model to contextualize local observations within a short-term history, analogous to how humans review recent temporal context to disambiguate intent and action quality.

\subsubsection{MIMO Transformer Architecture}
\label{subsec:mimo_arch}
We adopt the Transformer Sequential Aggregator from SARM~\cite{chen2025sarmstageawarerewardmodeling} as our backbone, re-engineering it to support multi-frame causal reasoning and relative advantage estimation.
ARM processes a sequence of historical observations within a causal window $\mathcal{W}_t = \{o_{t-4k}, \dots, o_t\}$ in parallel. 
By restricting the receptive field to past frames, this window-based approach ensures that predictions are informed by sufficient motion cues while maintaining real-time inference capabilities.
Crucially, this causal formulation ensures seamless compatibility with both online and offline RL paradigms, as it facilitates instantaneous reward generation without any dependency on future trajectory segments.

\paragraph{Multimodal Fusion.} 
For each timestep $i \in \mathcal{W}_t$, ARM integrates three disparate signals: (i) CLIP-based~\cite{radford2021learningtransferablevisualmodels} visual features $v_i \in \mathbb{R}^{d_{vis}}$, (ii) robot proprioceptive states $s_i \in \mathbb{R}^{d_{state}}$, and (iii) task instructions $g \in \mathbb{R}^{d_{lang}}$. These inputs are projected into a unified $d$-dimensional latent space to form a fused multimodal embedding $x_i$, defined as:
\begin{equation}
x_i = \text{MLP}(v_i) + \text{MLP}(s_i) + \text{MLP}(g)
\end{equation}
The resulting sequence $\{x_i\}_{i=t-4k}^t$ is then processed by an 8-layer Transformer Encoder to yield temporally enriched latent representations $\{h\}$:
\begin{equation}
\{h_{t-4k}, \dots, h_t\} = \text{Transformer} \left( \{ x_i \}_{i=t-4k}^{t} \right)
\end{equation}
where each $h_i$ encodes the historical evolution and kinematic state of the task at that specific moment.

\paragraph{Dual-Head Learning Objective.}
To balance sensitivity to local state transitions with the perception of global task goals, ARM is optimized via two synergistic output heads:

\begin{enumerate}
    \item \textbf{Multi-frame Advantage Classification:}
    The interval head infers the advantage transitions $\Delta \hat{y}$ between consecutive hidden states $(h_i, h_{i+1})$. This branch is optimized via a standard cross-entropy loss, denoted as $\mathcal{L}_{\text{int}}$, which is supervised by the tri-state labels (detailed in Sec.~\ref{labeling}). By reformulating reward estimation as a discrete classification task rather than continuous regression, the model exhibits significantly enhanced robustness against the non-linear noise and temporal stochasticity inherent in demonstrations.

    \item \textbf{Task Completion Prediction:}
    To anchor relative advantage estimations to absolute task metrics, the completion head $C$ predicts the probability that the current observation $s_t$ constitutes a successful terminal state. This decoupled design not only facilitates the identification of successful task executions, but also extracts progress anchor points from the predictions. When jointly utilized with the Multi-frame Advantage Classification results, these anchor points enable highly consistent, dense progress reconstructions.
    
    Moreover, since successful terminal frames are exceedingly rare within long-horizon continuous trajectories, this branch suffers from severe class imbalance. To effectively address this issue, we optimize the completion head using Focal Loss~\cite{lin2018focallossdenseobject}:
    \begin{equation}
    \mathcal{L}_{succ} = \text{FocalLoss}(C_t, \mathbb{1}[P_t \geq 1 - \epsilon])
    \end{equation}
\end{enumerate}

The total objective is defined as $\mathcal{L}_{ARM} = \lambda_{int} \mathcal{L}_{int} + \lambda_{succ} \mathcal{L}_{succ}$. This joint training enables the model to not only recover continuous progress curves but also accurately identify regressive behaviors and critical task completion moments.

\begin{figure*}[t]
    \centering
    \includegraphics[width=0.95\textwidth]{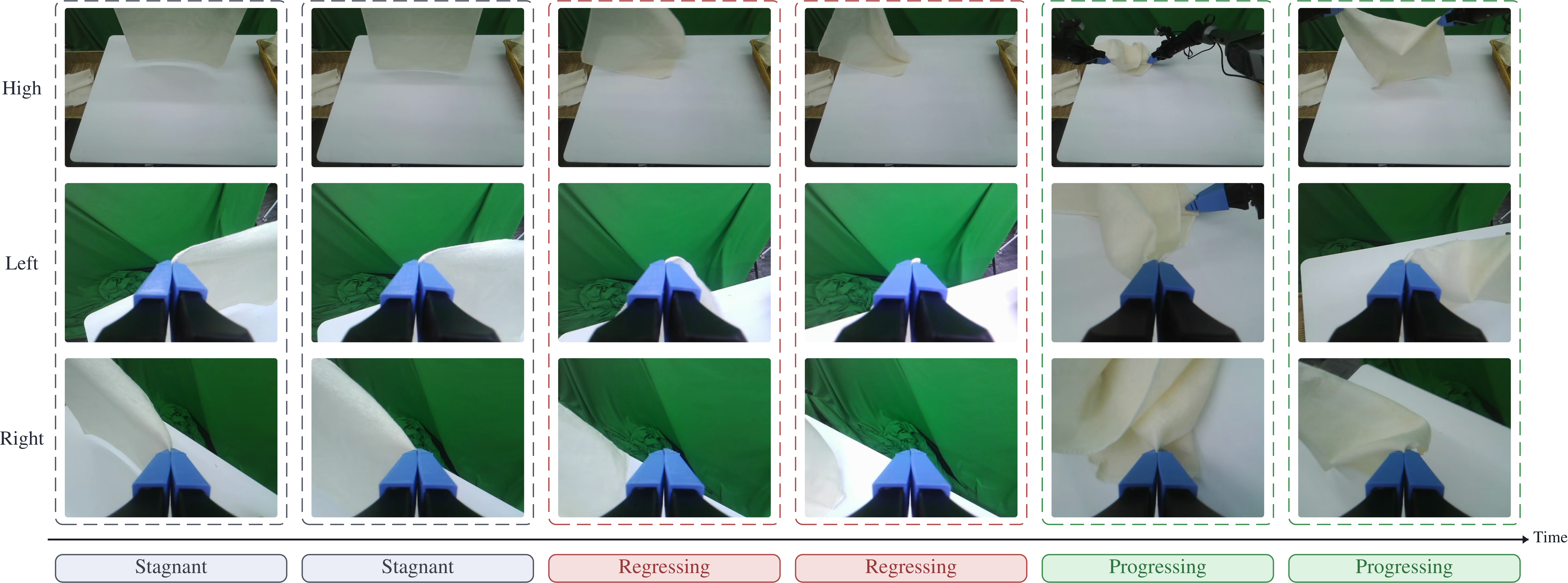}
    \caption{Illustration of the tri-state labeling strategy applied to a demonstration episode.}
    \label{fig:reward_layout}
\end{figure*}
\vspace{0.5em}
\subsubsection{Lightweight Tri-state Auto Labeling Strategy}
\label{labeling}

Traditional reward engineering for robotic manipulation typically requires annotators to assign a normalized scalar value \(P \in [0, 1]\) to each video frame. This continuous labeling process imposes a high cognitive load and is prone to inter-annotator inconsistency, as the definition of ``progress'' is often subjective. Such noise in supervision signals frequently leads to suboptimal policy convergence and substantial engineering overhead.

To address these issues, we redefine the annotation task as a \textbf{tri-state categorical classification of relative advantage}. 
As illustrated in Figure~\ref{fig:reward_layout}, for any observation pair \((s_t, s_{t+k})\), we define a progress-based advantage label \(y \in \{-1, 0, +1\}\) according to the following rules:

\begin{itemize}
    \item \textbf{+1 (Progressing):} The state effectively advances toward the task goal.
    \item \textbf{-1 (Regressing):} The state deviates from the goal, encounters an error, or results in failure.
    \item \textbf{0 (Stagnant):} No substantial progress is made, corresponding to waiting or idle behavior.
\end{itemize}

By acquiring initial human annotations through this simplified paradigm, we can efficiently cold-start our model. Subsequently, the trained model is utilized to perform inference on vast amounts of unannotated trajectories, automatically generating large-scale pseudo-labeled data for further training.

\vspace{0.5em}

\subsection{Global Progress Reconstruction}
\label{sec:global_reconstruction} 

As illustrated in Fig.~\ref{fig:arm_architecture}B, leveraging the \textbf{MIMO} architecture enables ARM to decompose complete video demonstrations and systematically aggregate the resulting predictions to reconstruct a dense, full-sequence progress curve:

\begin{enumerate}
    \item \textbf{Parallel Inference Efficiency:} While traditional sliding-window methods suffer from redundant computations on overlapping frames, the MIMO architecture predicts sequences directly within its context window. By leveraging video clipping, lengthy episodic trajectories are partitioned into independent, non-overlapping segments. These segments can be processed concurrently as parallel batches in a single forward pass, significantly accelerating the overall inference process.
    
    \item \textbf{Sequence Alignment and Padding:} For terminal video segments that are shorter than the model's specified window size, a tail-frame replication padding strategy is applied. During the final aggregation of the full episode, predictions corresponding to these synthetically padded regions are discarded to maintain temporal fidelity.
    
    \item \textbf{Coherent Progress Generation:} To generate the global dense progress curve $P_t$, the system mathematically integrates the model-predicted relative state transitions $\Delta \hat{y}$ with the absolute task completion signal $C_t$. Specifically, treating $C_t$ as the definitive progress anchor (e.g., $P_T = 1.0$ at task completion), the dense progress values for preceding frames are reconstructed via accumulation of $\Delta \hat{y}$.
\end{enumerate}

This pipeline elegantly transforms discrete, local relative predictions into a coherent and dense global progress signal, thereby providing consistent, high-quality supervision for subsequent policy learning.

\subsection{Policy Optimization via AW-BC}

Based on the dense progress signals reconstructed by ARM, we propose \textbf{Advantage-Weighted Behavior Cloning (AW-BC)}. This framework prioritizes learning from high-advantage transitions while suppressing suboptimal behaviors through a statistically grounded reweighting mechanism, as illustrated in Fig.~\ref{fig:arm_architecture}C.

\subsubsection{Length-adaptive Gain Formulation}
To mitigate the length bias inherent in heterogeneous demonstrations—where drastic variations in episode duration lead to inconsistent progress gradients (e.g., disproportionately steep slopes in shorter sequences)—we introduce an adaptive scaling mechanism. Such gradient volatility often induces instability and jitter in the learning dynamics, hindering smooth weight optimization. For an action chunk with horizon $H$, the \textbf{length-adaptive gain} $\Delta G_t$ is formulated as:

\begin{equation}
\Delta G_t = (P_{t+H} - P_t) \cdot \frac{L_{\text{seq}}}{\bar{L}}
\end{equation}

where $P_t$ denotes the progress value obtained via global progress reconstruction, $L_{\text{seq}}$ represents the total length of the current episode, and $\bar{L}$ is the average episode length across the entire dataset. 
This normalization ensures that the derived advantage reflects the relative efficiency of a specific action sequence, effectively decoupling the reward signal from the absolute duration of the task. 
\subsubsection{Statistical Weighting and Objective}
To convert raw gains into robust training weights, we employ a statistical normalization strategy based on the gain distribution of the current batch. Let $\mu$ and $\sigma$ be the mean and standard deviation of $\{\Delta G_i\}$. We define clipping bounds as $b_{lower} = \mu - 2\sigma$ and $b_{upper} = \mu + 2\sigma$. The importance weight $\tilde{w}_i$ for each sample is computed as:

\begin{equation}
\tilde{w}_i = \text{clamp}\left( \frac{\Delta G_i - b_{lower}}{b_{upper} - b_{lower} + \epsilon}, \ 0, \ 1 \right)
\end{equation}

This clamping mechanism effectively filters out regressive data (weights $\to 0$) while capping the influence of outliers. The final AW-BC objective is to minimize the weighted negative log-likelihood:

\begin{equation}
\mathcal{L}_{AW-BC}(\theta) = \mathbb{E}_{(s, a) \sim \mathcal{D}} \left[ - \tilde{w}(s,a) \log \pi_{\theta}(a|s) \right]
\end{equation}

\subsubsection{Theoretical Connection to Offline RL}
Our proposed formulation aligns with the principles of \textbf{AWR}~\cite{peng2019advantageweightedregressionsimplescalable}. Mathematically, this optimization problem can be viewed as maximizing the expected return of the policy under the constraint of remaining close to the behavior policy:

\begin{equation}
\max_{\theta} \mathbb{E}_{(s,a) \sim \mathcal{D}} \left[ \tilde{w}(s,a) \log \pi_{\theta}(a|s) \right]
\end{equation}

Here, ARM functions as a learned \textbf{Critic}, providing the advantage estimate $\Delta G_t$ that guides the policy update. By prioritizing transitions with high relative advantage, our method effectively performs offline policy improvement, extracting an optimal policy from suboptimal demonstrations without explicit online interaction.

%% file: sec/4_exp.tex
\section{Experiments}
\label{sec:experiments}

\subsection{Experimental Setup}
\label{sec:experimental_setup}

\begin{figure}[htbp]
    \centering
    \includegraphics[width=1.0\linewidth]{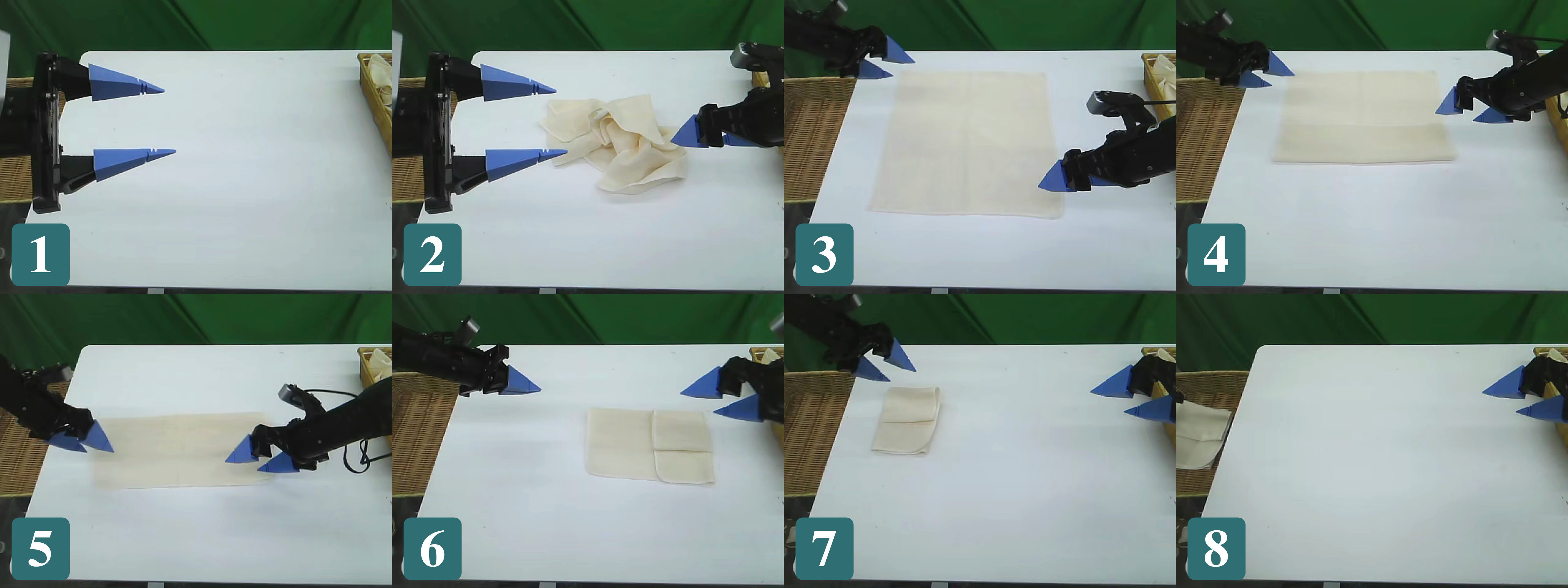}
    \caption{Overview of the long-horizon towel-folding task. The sequence includes extracting a towel from clutter, placing and flattening it on the table, executing a precise multi-stage folding strategy, and transporting the folded towel into the target box.}
    \label{fig:task_setup}
\end{figure}

We evaluate our framework on a challenging, long-horizon bimanual towel-folding task. As illustrated in Fig.~\ref{fig:task_setup}, a complete and successful demonstration requires a structured 8-stage procedure: 
(1) \textbf{extracting} exactly one towel from an unstructured, cluttered pile; 
(2) \textbf{placing} it onto the central tabletop; 
(3) \textbf{flattening} the towel to a planar initial state; 
(4) \textbf{performing} a bottom-to-up longitudinal fold; 
(5) \textbf{executing} a top-to-bottom longitudinal fold; 
(6) \textbf{conducting} a right-to-center lateral fold; 
(7) \textbf{completing} the sequence with a left-to-right lateral fold to form a compact rectangle; and 
(8) \textbf{transporting} and depositing the folded towel fully inside a target storage box on the left. 
A trial is considered successful only if a single towel is extracted, remains neatly folded, and is fully contained within the box boundaries within a 120-second limit. 
\paragraph{Task and Hardware.} 
Data was collected using an AgileX ALOHA~\cite{fu2024mobile} bimanual teleoperation system with randomized table heights for enhanced generalization. Detailed implementation details are in the Supplementary Materials.

\paragraph{Dataset Construction and Labeling.} 
We curated a dataset $\mathcal{D}_{all}$ of 972 towel-folding episodes (20 hours total), comprising 809 expert demonstrations and 163 DAgger-augmented error-correction episodes. Unlike SARM~\cite{chen2025sarmstageawarerewardmodeling}, we retain all trajectories including slow episodes that contain important recovery patterns.

We evaluate three annotation paradigms: (i) \textbf{VLM-based Labeling} implemented in LeRobot~\cite{lerobot}, using Qwen3-VL~\cite{QwenLM_Qwen3_VL} for temporal grounding of subtask boundaries; (ii) \textbf{Manual Subtask Segmentation} by human experts; and (iii) our proposed \textbf{Tri-state Labeling}. 

\subsection{Reward Model Performance}
\label{sec:reward_model_performance}

To systematically evaluate the precision and robustness of our proposed reward models, we compare \textbf{ARM} and \textbf{SARM} ~\cite{chen2025sarmstageawarerewardmodeling} against the Ground Truth (GT). The evaluation metrics focus on two primary aspects: the numerical accuracy of progress estimation (MSE) and the categorical reliability of trajectory classification.

\paragraph{Quantitative Results.} 
Table~\ref{tab:reward_performance} summarizes the quantitative results. As expected, ARM demonstrates superior alignment with the GT signals across all evaluation criteria compared to SARM. Notably, ARM achieves a significantly lower MSE (0.0014 vs. 0.0059), representing a substantial improvement in the fidelity of dense progress estimation. Furthermore, ARM achieves perfect success rates in identifying Standard (SE) and Failure (FE) episodes, underscoring its robustness in diverse terminal scenarios.

\begin{table}[t]
\centering
\caption{\textbf{Quantitative Evaluation of Reward Models.} 
All models are evaluated on a validation set of \textbf{50 trajectories}. 
``MSE'' measures the trajectory reconstruction fidelity against GT progress (normalized to $[0, 1]$). 
The bottom section reports the \textbf{Success Identification Accuracy}, assessing the Completion Head's ability to correctly classify the final state of Standard (\textbf{SE}, 12 successful episodes), and Failure (\textbf{FE}, 12 failed episodes) trajectories. 
Best performances are highlighted in \textbf{bold}.}
\label{tab:reward_performance}

\vspace{1mm}
\small
\renewcommand{\arraystretch}{1.03}
\setlength{\tabcolsep}{5pt}

\begin{tabularx}{\columnwidth}{
>{\centering\arraybackslash}X
>{\centering\arraybackslash}X
>{\centering\arraybackslash}X}

\specialrule{1pt}{0pt}{1pt}
\toprule

\textbf{Metrics}
&
\textbf{SARM}
&
\textbf{ARM (Ours)}

\\

\midrule

MSE $\downarrow$
&
0.0059
&
\textbf{0.0014}

\\

\midrule

\multicolumn{3}{c}{\textit{Success Identification Accuracy (\%)}}

\\

Standard (SE)
&
83.3 (10/12)
&
\textbf{100.0 (12/12)}

\\

\rowcolor[gray]{.9}
Failure (FE)
&
91.6 (11/12)
&
\textbf{100.0 (12/12)}

\\

\bottomrule
\specialrule{1pt}{1pt}{0pt}

\end{tabularx}

\vspace{1mm}

\end{table}

\begin{figure}[htbp]
    \centering
    \includegraphics[width=1.0\linewidth]{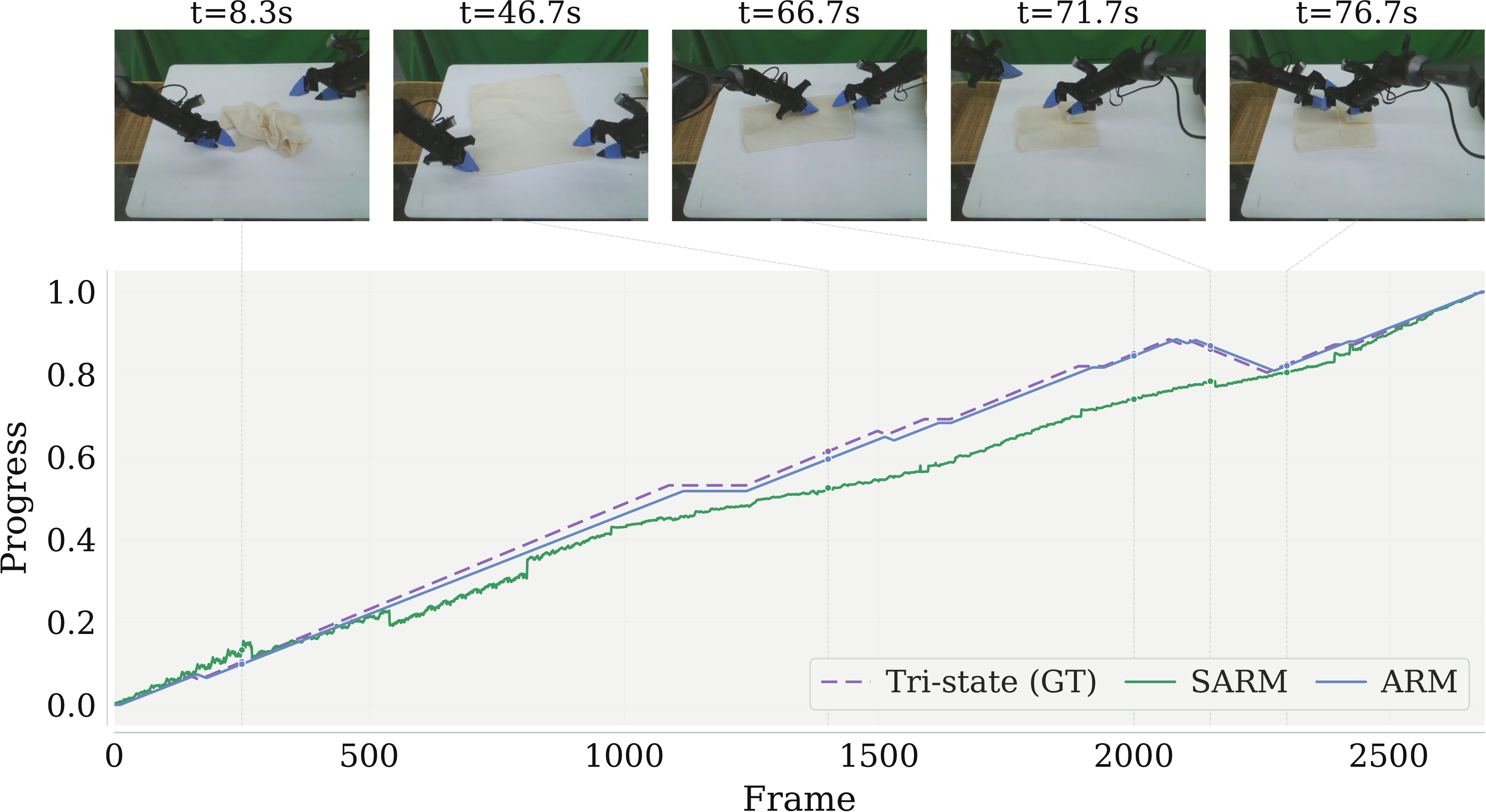}
    \caption{\textbf{Qualitative comparison of progress reconstruction.} We visualize the progress curves of SARM and ARM against the GT for a representative episode. While SARM struggles with non-monotonic behaviors, ARM reconstructs a smooth, high-fidelity curve that closely tracks the GT, even during regressive adjustments.}
    \label{fig:progress_comparison}
\end{figure}

\paragraph{Qualitative Analysis.} 
Fig.~\ref{fig:progress_comparison} visualizes the progress reconstruction differences between SARM and ARM. \textbf{SARM} produces stepped curves with abrupt transitions at subtask boundaries, failing to capture localized regressive movements. In contrast, \textbf{ARM} leverages relative advantage signals to generate smooth, dense progress curves that closely track the ground truth, even during non-monotonic robot adjustments.

\begin{table*}[t]
\centering
\caption{\textbf{Quantitative Comparison of Downstream Policy Performance.} 
We report the success rate, operational task throughput (episodes completed per hour), and folding precision (final edge alignment score; detailed annotation protocol provided in the Supplementary Material) on the long-horizon towel-folding task. Our proposed \textbf{AW-BC (ARM)} framework significantly outperforms both standard Behavior Cloning and prior reward-aware baselines across all metrics.}
\label{tab:policy_performance}

\vspace{1mm}
\small
\renewcommand{\arraystretch}{1.05}
\setlength{\tabcolsep}{5pt}

\begin{tabularx}{\textwidth}{
>{\centering\arraybackslash}X
!{\vrule width 0.6pt}
>{\centering\arraybackslash}X
>{\centering\arraybackslash}X
>{\centering\arraybackslash}X
}

\specialrule{1pt}{0pt}{1pt}
\toprule

\textbf{Method}
&
\textbf{Success Rate}
&
\textbf{Task Throughput}
&
\textbf{Folding Precision}
\\

&
\small{(\%)}
&
\small{(Episodes / hr)}
&
\small{(Score)}
\\

\midrule

BC-Baseline (GR00T N1.5)
&
62.1
&
18
&
2.2
\\

RA-BC (GR00T + SARM)
&
78.5
&
24
&
2.7
\\

\rowcolor[gray]{.9}
\textbf{AW-BC (GR00T + ARM)}
&
\textbf{99.4}
&
\textbf{32}
&
\textbf{3.6}
\\

\bottomrule
\specialrule{1pt}{1pt}{0pt}

\end{tabularx}

\vspace{1mm}

\end{table*}

\subsection{Efficiency and Quality of Reward Labeling}
\label{sec:labeling_analysis}

\begin{figure}[t]
    \centering
    \includegraphics[width=1.0\linewidth]{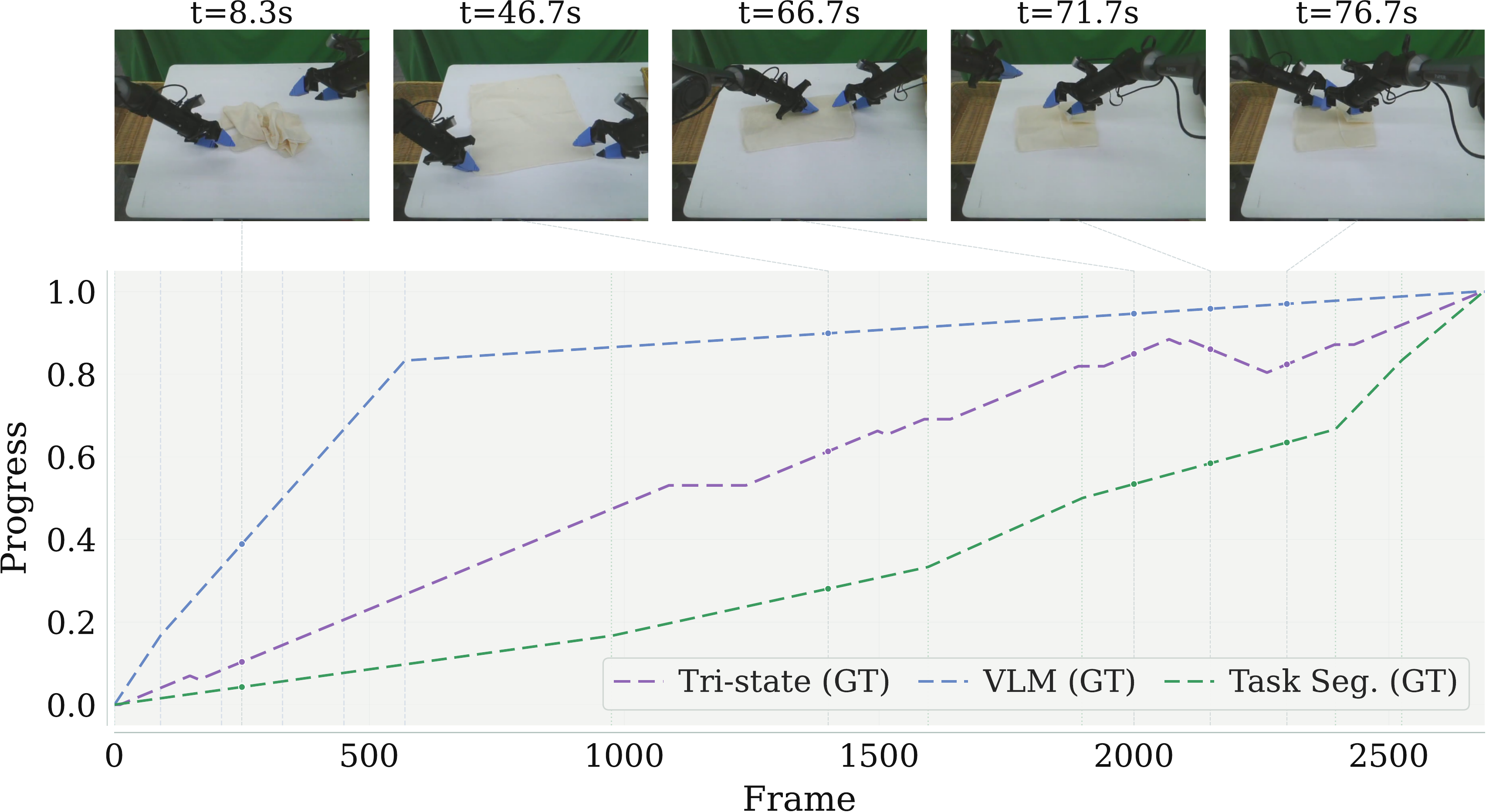}
    \caption{Qualitative comparison of progress reconstruction. Our tri-state approach generates smoother, more consistent dense progress signals compared to the stepped curves of manual segmentation and VLM methods.}
    \label{fig:labeling_comparison}
\end{figure}

A primary bottleneck in scaling reward-guided behavior cloning is the prohibitive cost of human annotation. To evaluate our framework, we conducted a controlled user study with five annotators comparing our \textbf{Tri-state Advantage Labeling} against the \textbf{Subtask Segmentation} protocol (visualized in Fig.~\ref{fig:labeling_comparison}). We evaluate the labeling process from two dimensions: throughput efficiency and reconstruction quality.

\begin{table}[t]
\centering
 \caption{\textbf{Labeling Efficiency Comparison.} Annotation throughput comparison between human and automated labeling protocols per 8-hour
  shift.}
\label{tab:efficiency_performance}

\small
\setlength{\tabcolsep}{6pt}

\begin{threeparttable}

\begin{tabularx}{\columnwidth}{>{\centering\arraybackslash}X >{\centering\arraybackslash}X}

\specialrule{1pt}{0pt}{1pt}
\toprule

\textbf{Annotation Protocol} 
& 
\textbf{Rate (Samples/8h)} \\

\midrule

Human Baseline (Seg.)\tnote{$\dagger$}
&
100 \\

\rowcolor[gray]{.9}
\textbf{Human Tri-state (Ours)}\tnote{$\dagger$}
&
\textbf{250} \\

\midrule

VLM (Qwen3-VL)\tnote{$\ddagger$}
&
$\sim 3000$ \\

\rowcolor[gray]{.9}
\textbf{Auto Tri-state (Ours)}\tnote{$\ddagger$}
&
$\mathbf{>400,000}$ \\

\bottomrule
\specialrule{1pt}{1pt}{0pt}

\end{tabularx}

\begin{tablenotes}[flushleft]
\scriptsize
\item[$\dagger$] Per single human annotator.
\item[$\ddagger$] Inference throughput on a single NVIDIA A100 GPU.
\end{tablenotes}

\end{threeparttable}

\end{table}

\paragraph{Labeling Throughput and Quality.}
As shown in Table~\ref{tab:efficiency_performance}, our tri-state protocol achieves significant efficiency gains. By simplifying annotation from precise temporal boundary localization to discrete classification, human annotators achieve \textbf{250 samples} per 8-hour shift—a \textbf{2.5× speedup} over the baseline (100 samples). This simplified formulation enables massive scaling: our \textbf{Auto Tri-state} pipeline processes $>400,000$ samples per 8 hours, achieving a \textbf{$>133\times$ speedup} over human baselines.

Beyond efficiency, our approach provides superior signal quality. As shown in Fig.~\ref{fig:labeling_comparison}, manual and VLM methods produce stepped progress curves with temporal misalignment, while our tri-state labeling ($+1, 0, -1$) yields smooth, dense progress curves when integrated with ARM's anchor points.

\subsection{MIMO Architecture Efficiency Analysis}
\label{sec:mimo_efficiency}

To demonstrate the efficiency advantages of our proposed Multiple-Input Multiple-Output (MIMO) architecture, we conduct an ablation study comparing inference speeds across three distinct approaches: our ARM with MIMO design, traditional MISO VLM labeling using Qwen3-VL, and the baseline SARM model. The results, summarized in Table~\ref{tab:mimo_efficiency}, highlight the substantial computational benefits achieved through our architectural design.

\begin{table}[t]
\centering
\caption{\textbf{MIMO Architecture Efficiency Comparison.} We evaluate the inference throughput across different reward modeling approaches. ARM is evaluated with its MIMO design handling 5 parallel outputs per input, VLM labeling represents traditional single-input approaches, and SARM serves as the baseline. All measurements are conducted on a single NVIDIA A100 GPU under comparable conditions.}
\label{tab:mimo_efficiency}

\vspace{1mm}
\small
\renewcommand{\arraystretch}{1.05}
\setlength{\tabcolsep}{8pt}

\begin{tabularx}{\columnwidth}{
>{\raggedright\arraybackslash}X
>{\centering\arraybackslash}X
>{\centering\arraybackslash}X}

\specialrule{1pt}{0pt}{1pt}
\toprule

\textbf{Method}
&
\textbf{Architecture}
&
\textbf{Throughput (it/s)}
\\

\midrule

Qwen3-VL
&
MISO
&
1.03
\\

SARM Baseline
&
SISO
&
3.9
\\

\rowcolor[gray]{.9}
\textbf{ARM (Ours)}
&
\textbf{MIMO}
&
\textbf{14.1}
\\

\bottomrule
\specialrule{1pt}{1pt}{0pt}

\end{tabularx}

\vspace{1mm}

\end{table}

As demonstrated in Table~\ref{tab:mimo_efficiency}, our ARM achieves an inference speed of \textbf{14.1 iterations per second} (calculated as $2.82 \times 5$ for the 5-output MIMO configuration), representing a \textbf{13.7×} speedup over VLM-based labeling (1.03 it/s) and a \textbf{3.6×} improvement over SARM (3.9 it/s). This substantial efficiency gain stems from the MIMO architecture's ability to process multiple advantage predictions simultaneously within a single forward pass, eliminating the computational redundancy inherent in sequential processing approaches.

The efficiency advantage becomes particularly crucial during large-scale deployment, where the ARM model must process extensive trajectory datasets for reward signal generation. While traditional VLM approaches suffer from the overhead of processing each temporal segment independently, our MIMO design leverages shared feature representations to amortize computational costs across multiple outputs, making it highly scalable for real-world robotic learning applications.

\subsection{Policy Performance Analysis}
\label{sec:policy_performance}

We evaluate the downstream manipulation performance by comparing three distinct policy configurations based on the GR00T-N1.5-3B~\cite{gr00tn1}:
\begin{itemize}
    \item \textbf{(1) Baseline:} Standard Behavior Cloning trained on the full dataset $\mathcal{D}_{all}$.
    \item \textbf{(2) RA-BC (GR00T+SARM):} Reward-Aligned Behavior Cloning~\cite{chen2025sarmstageawarerewardmodeling} re-weighted by SARM progress signals.
    \item \textbf{(3) AW-BC (GR00T+ARM, Ours):} Our proposed policy trained via Advantage-Weighted Behavior Cloning, utilizing the dense, relative advantage signals from ARM.
\end{itemize}

As summarized in Table~\ref{tab:policy_performance}, the \textbf{Baseline} suffers from a suboptimal success rate (62.1\%) and lower operational efficiency. This is primarily due to the multi-modal noise and ``sluggish'' trajectories inherent in the full dataset, which typical BC fails to filter or prioritize. While \textbf{RA-BC(GR00T+SARM)} improves the success rate to 78.5\% through subtask-based weighting, it remains constrained by the lack of fine-grained advantage estimation for error-correction behaviors.

Crucially, our framework achieves a near-perfect success rate of \textbf{99.4\%}. Beyond reliability, our policy demonstrates superior \textbf{Task Throughput} (32 episodes/hr), significantly outperforming the baselines. This indicates that the advantage-weighted objective effectively prioritizes high-quality, decisive movements, resulting in more agile and purposeful trajectories. Furthermore, our method achieves the highest \textbf{Folding Precision} (3.6), as the dense reward signal provides finer supervision for the critical multi-stage alignment required in towel folding.

\paragraph{Ablation Study.}
To isolate the contributions of our key innovations, we evaluate three configurations through pairwise comparisons, as shown in Table~\ref{tab:ablation_study}.

\begin{table}[t]
\centering
\caption{\textbf{Ablation Study.}
We systematically evaluate the contributions of tri-state labeling and AW-BC training through three key configurations.}
\label{tab:ablation_study}

\vspace{1mm}
\small
\renewcommand{\arraystretch}{1.05}
\setlength{\tabcolsep}{6pt}

\begin{tabularx}{\columnwidth}{
>{\raggedright\arraybackslash}X
>{\centering\arraybackslash}X
>{\centering\arraybackslash}X
>{\centering\arraybackslash}X
>{\centering\arraybackslash}X
!{\vrule width 0.6pt}
>{\centering\arraybackslash}X
}

\specialrule{1pt}{0pt}{1pt}
\toprule

\textbf{Method}
&
\textbf{Task Seg.}
&
\textbf{Tri-state}
&
\textbf{RA-BC}
&
\textbf{AW-BC}
&
\textbf{Success Rate (\%)}
\\

\midrule

SARM
&
\checkmark
&
--
&
\checkmark
&
--
&
78.5
\\

ARM 
&
--
&
\checkmark
&
\checkmark
&
--
&
92.3
\\

\rowcolor[gray]{.9}
\textbf{ARM}
&
--
&
\checkmark
&
--
&
\checkmark
&
\textbf{99.4}
\\

\bottomrule
\specialrule{1pt}{1pt}{0pt}

\end{tabularx}

\vspace{1mm}

\end{table}

The results reveal the impact of each component through direct comparisons.

\textbf{Tri-state vs. Task Segmentation}: Comparing SARM with ARM (Tri-state + RA-BC) shows tri-state labeling improves success rate from 78.5\% to 92.3\% (+13.8\%), demonstrating superior annotation quality and efficiency.

\textbf{AW-BC vs. RA-BC}: Comparing ARM (Tri-state + RA-BC) with ARM (Tri-state + AW-BC) shows our advantage-weighted training dramatically improves success rate from 92.3\% to 99.4\% (+7.1\%), highlighting the effectiveness of dense advantage signals.

Our complete ARM framework achieves +20.9\% improvement over SARM, demonstrating strong synergy between tri-state labeling and AW-BC training.

%% file: sec/5_conclusion.tex
\section{Conclusion}
We propose \textbf{Advantage Reward Model (ARM)}, a framework that addresses the reward engineering bottleneck in long-horizon robotic manipulation tasks. By modeling relative advantages, ARM overcomes inconsistency and high costs of traditional dense labeling. We introduce a \textit{tri-state labeling strategy} that reduces cognitive load for annotators while providing high-fidelity supervision signals and enabling automated labeling. In a challenging towel-folding task, ARM with Advantage-Weighted Behavior Cloning achieves a $99.4\%$ success rate, outperforming existing Vision-Language-Action baselines. ARM provides a scalable and robust solution for training high-performance policies.

%% file: sec/X_suppl.tex
\clearpage 
\appendix

\section*{Author Contributions}

\noindent \textbf{Yiming Mao} is the primary architect of the ARM framework and spearheaded its development from the ground up. He designed the core algorithms, performed comprehensive hardware-software debugging. He conducted the entirety of the robotic manipulation experiments, managed the complete data engineering workflow, and drafted the original manuscript.\smallskip

\noindent \textbf{Zixi Yu} contributed to manuscript drafting, prepared the technical illustrations and figures, and assisted in the replication of baseline methods. \smallskip

\noindent \textbf{Weixin Mao} served as the Project Leader, providing overall supervision and strategic steering of the research direction. He played a key role in the intellectual refinement of the framework and critically revised the manuscript to ensure its technical and academic rigor. \smallskip

\noindent \textbf{Yinhao Li} provided the initial software infrastructure and codebase. \smallskip

\noindent \textbf{Qirui Hu} assisted with the maintenance and debugging of the robot hardware. \smallskip

\noindent \textbf{Zihan Lan} contributed to the data parsing scripts. \smallskip

\noindent \textbf{Minzhao Zhu} participated in technical discussions and provided general support. \smallskip

\noindent \textbf{Hua Chen} provided administrative support and coordinated the research resources.

\section{VLM Prompting Details}


For the towel-folding task, the subtask vocabulary is:
\begin{enumerate}
\item Extracting exactly one towel from an unstructured, cluttered pile;
\item Placing it onto the central tabletop;
\item Flattening the towel to a planar initial state;
\item Performing a bottom-to-up longitudinal fold;
\item Executing a top-to-bottom longitudinal fold;
\item Conducting a right-to-center lateral fold;
\item Completing the sequence with a left-to-right lateral fold to form a compact rectangle;
\item Transporting and depositing the folded towel fully inside a target storage box on the left.
\end{enumerate}

The effective prompt is:

\begingroup
\ttfamily
\# Role\par

You are a Robotics Vision System specializing in temporal action localization
for robot manipulation. Your job is to segment a single demonstration video
into distinct, non-overlapping atomic actions from a fixed label list.\par

\# Label Set (Closed Vocabulary)\par

You must strictly identify the video segments using ONLY the provided labels.
Do not create new labels or modify existing ones.\par

The video shows execution of all actions in logical orders.\par

\# Ground-Truth Semantics\par

Use visual state changes to define when an action starts and ends. Do NOT
assume equal durations for the stages.\par

- An action starts at the first frame where the robot's motion clearly
initiates that action.\par

- An action ends at the first frame where that specific action is visually
completed and the manipulated object reaches a temporary, stable
configuration.\par

- Short pauses or ambiguous micro-motions should be assigned to the current
action.\par

\# Constraints\par

1. The full video from ``00:00'' to the final timestamp must be covered
without gaps.\par

2. The end timestamp of one stage must equal the start timestamp of the next
stage.\par

3. Each stage appears exactly once and in logical order.\par

4. Uniform or near-uniform segmentation should be avoided unless the video
genuinely supports it.\par

5. Timestamps must be in ``MM:SS'' format; the first stage starts at
``00:00''.\par

\# Step 1 --- Textual Timeline\par

First, write a detailed textual timeline with approximate timestamps. For each
stage, include its name, approximate start and end time, and the visual event
that defines the boundary.\par

\# Step 2 --- Structured Output\par

Then output only valid JSON consistent with the timeline above, using the
exact labels and timestamps without adding extra keys.\par
\endgroup

In the implementation, this prompt is provided as a system instruction, while the user message contains the episode video and a short duration hint formatted as ``Video is MM:SS ($\sim$xx.xs). Follow instructions.'' The resulting VLM output is parsed into stage names with start and end timestamps and then written into the dense subtask annotation fields of the dataset.

\section{Implementation Details}
\label{sec:implementation_details}

Our framework consists of two primary components: the \textbf{Advantage Reward Model (ARM)} and the \textbf{Policy Model}, both of which leverage high-capacity pre-trained backbones but are optimized for distinct objectives.

\paragraph{Reward Model (ARM) Training.}
ARM utilizes a pre-trained \textbf{CLIP ViT-B/32} as the vision-text encoder, followed by a Transformer-based Sequential Aggregator with a causal 5-frame window (sampled at 1Hz). The joint objective is defined as $\mathcal{L}_{\text{ARM}} = \lambda_{\text{int}} \mathcal{L}_{\text{int}} + \lambda_{\text{succ}} \mathcal{L}_{\text{succ}}$, where we employ Focal Loss for task completion and cross-entropy for tri-state interval classification. Complete hyperparameters are summarized in Table~\ref{tab:arm_hyperparams}.

\paragraph{Policy Training (AW-BC).}
Based on the \textbf{GR00T-N1.5} VLA foundation, our policy uses Advantage-Weighted Behavior Cloning where sample weights $w$ are derived from ARM-predicted gains $\Delta G_t$. Training configurations are detailed in Table~\ref{tab:policy_hyperparams}.

\begin{table}[htbp]
\centering
\caption{\textbf{ARM Training Hyperparameters.} Complete hyperparameter settings for training the Advantage Reward Model.}
\label{tab:arm_hyperparams}

\small
\renewcommand{\arraystretch}{1.1}
\setlength{\tabcolsep}{8pt}

\begin{tabularx}{\columnwidth}{
>{\raggedright\arraybackslash}X
>{\centering\arraybackslash}X}

\specialrule{1pt}{0pt}{1pt}
\toprule

\textbf{Parameter}
&
\textbf{Value}
\\

\midrule

Vision Encoder
&
CLIP ViT-B/32
\\

Sequential Aggregator Window
&
5 frames (1Hz sampling)
\\

Training Epochs
&
2
\\

Hardware Configuration
&
2 × NVIDIA A100 GPUs
\\

Effective Batch Size
&
64
\\

\midrule

\multicolumn{2}{c}{\textit{Optimization Configuration}}
\\

Optimizer
&
AdamW
\\

Learning Rate (LR)
&
$5 \times 10^{-5}$
\\

Weight Decay (WD)
&
$10^{-3}$
\\

LR Warmup Steps
&
1,000
\\

LR Schedule
&
Cosine Decay
\\

Mixed Precision
&
FP16
\\

\midrule

\multicolumn{2}{c}{\textit{Loss Function Configuration}}
\\

Interval Loss Weight ($\lambda_{\text{int}}$)
&
1.0
\\

Success Loss Weight ($\lambda_{\text{succ}}$)
&
1.0
\\

Focal Loss $\gamma$
&
2.0
\\

Focal Loss $\alpha$
&
2.0
\\

Focal Loss $\epsilon$
&
$10^{-3}$
\\

\bottomrule
\specialrule{1pt}{1pt}{0pt}

\end{tabularx}

\end{table}

\begin{table}[htbp]
\centering
\caption{\textbf{Policy Training Hyperparameters.} Complete hyperparameter settings for Advantage-Weighted Behavior Cloning using the GR00T-N1.5 foundation model.}
\label{tab:policy_hyperparams}

\small
\renewcommand{\arraystretch}{1.1}
\setlength{\tabcolsep}{8pt}

\begin{tabularx}{\columnwidth}{
>{\raggedright\arraybackslash}X
>{\centering\arraybackslash}X}

\specialrule{1pt}{0pt}{1pt}
\toprule

\textbf{Parameter}
&
\textbf{Value}
\\

\midrule

Foundation Model
&
GR00T-N1.5 (3B parameters)
\\

Policy Head
&
Diffusion Transformer (DiT) Flow Matching
\\

Action Dimension
&
14D bimanual actions
\\

Action Horizon ($H$)
&
32
\\

Camera Views
&
3 × $224 \times 224$ (head + wrists)
\\

Training Epochs
&
7
\\

Hardware Configuration
&
32 × NVIDIA A100 GPUs
\\

Parallelization Strategy
&
FSDP (Fully Sharded Data Parallel)
\\

\midrule

\multicolumn{2}{c}{\textit{Optimization Configuration}}
\\

Batch Size
&
256
\\

Learning Rate
&
$2 \times 10^{-5}$ (constant)
\\

Mixed Precision
&
BF16
\\

Gradient Clipping
&
1.0
\\

\midrule

\multicolumn{2}{c}{\textit{Advantage Weighting Configuration}}
\\

Weight Clipping Range
&
$[0, 1]$
\\

Positive Threshold ($\Delta G_t > 0.01$)
&
$w = 1$
\\

Non-positive Threshold ($\Delta G_t \leq 0$)
&
$w = 0$
\\

\midrule

\multicolumn{2}{c}{\textit{Inference Configuration}}
\\

Flow Matching Denoising Steps
&
4
\\

\bottomrule
\specialrule{1pt}{1pt}{0pt}

\end{tabularx}

\end{table}

\section{ARM Inference Results}
\label{sec:arm_inference_results}

\begin{figure}[htbp]
    \centering
    \includegraphics[width=1.0\linewidth]{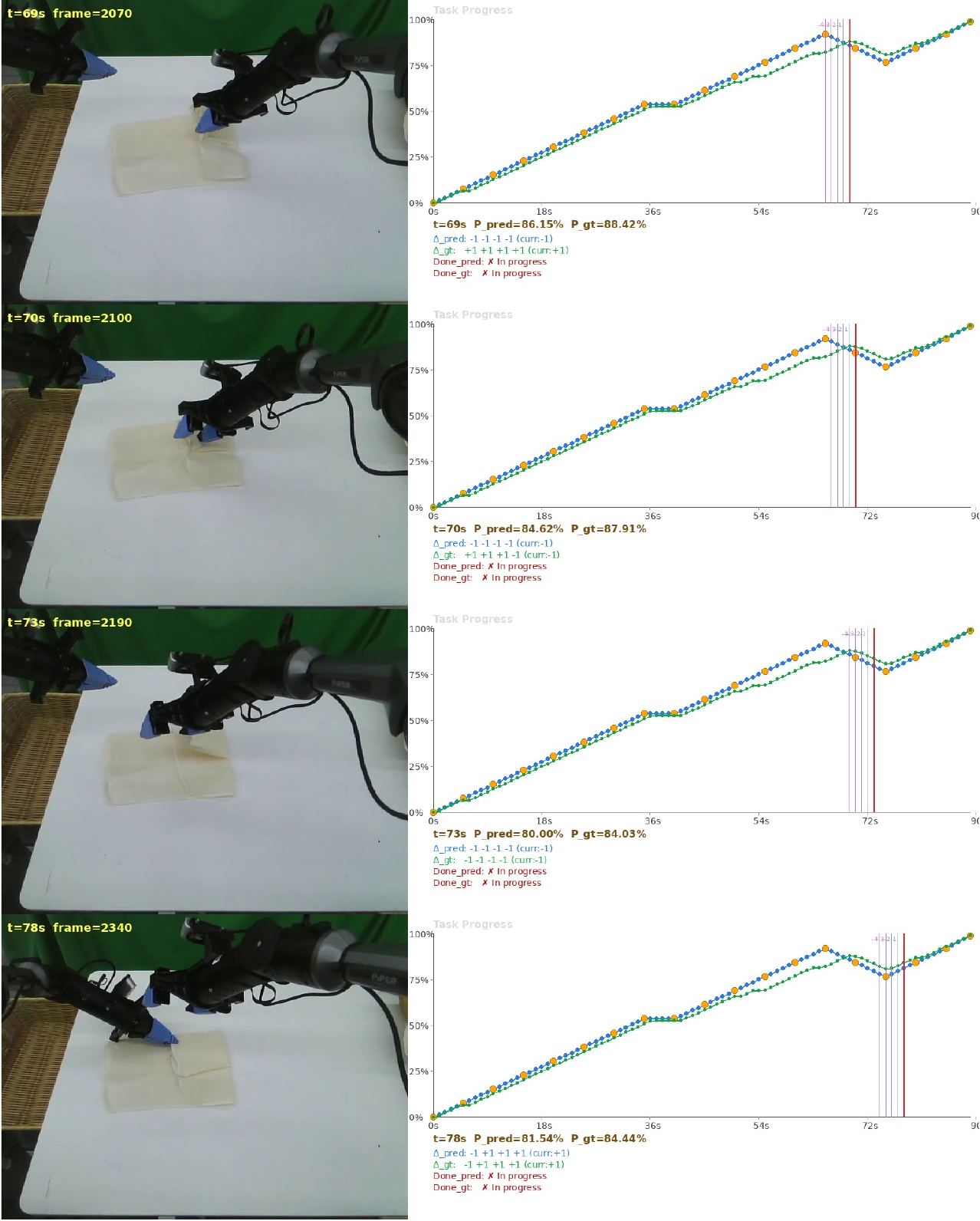}
    \caption{\textbf{Visualization of ARM Inference Results.} The left panels show the third-person view of the bimanual towel-folding task at $t=69s$ and $t=70s$. The right panels display the corresponding progress curves: predicted progress $P_{pred}$ (blue) and ground truth $P_{gt}$ (green). ARM accurately captures the non-monotonic progress ``dip'' caused by a regressive adjustment, with the Multi-frame Advantage head correctly outputting $\Delta_{\textbf{pred}} = -1$.}
    \label{fig:reward_infer_example}
\end{figure}

To evaluate the qualitative performance of our model, we visualize the \textbf{ARM inference results} on a held-out test trajectory, as shown in Fig.~\ref{fig:reward_infer_example}. The model is required to reconstruct a dense progress signal for a long-horizon towel-folding sequence characterized by non-monotonic behaviors.

\paragraph{Tracking Regressive Behaviors.} 
A critical observation in the inference results is ARM's sensitivity to physical regressions. Between $t=65s$ and $t=75s$, the robot performs a localized adjustment of the towel's edge to prepare for the final fold. This action, while necessary, temporarily moves the state further from the target rectangular configuration. 

As captured in the transition from $t=69$s ($P_{\text{pred}}=86.15\%$) to $t=70$s ($P_{\text{pred}}=84.62\%$), the Multi-frame
Advantage head successfully identifies this trend, consistently predicting regressive signals ($\Delta_{\textbf{pred}} = -1$, as shown in the status text). This causes the reconstructed progress curve (blue line) to exhibit a precise downward ``dip'' that closely aligns with the ground truth (green line). 

\paragraph{High-Fidelity Signal Reconstruction.} 
Despite the complexity of the 14-dimensional bimanual action space and the deformable nature of the towel, ARM maintains high temporal consistency throughout the inference process. The predicted curve is smooth and free from the cumulative drift or ``stepped'' artifacts common in subtask-based approaches. This high-fidelity inference result demonstrates that ARM can provide the downstream policy with accurate, real-time feedback, penalizing regressive movements and rewarding only those that effectively contribute to task completion.
\begin{figure}[htbp]
    \centering
    \includegraphics[width=0.95\linewidth]{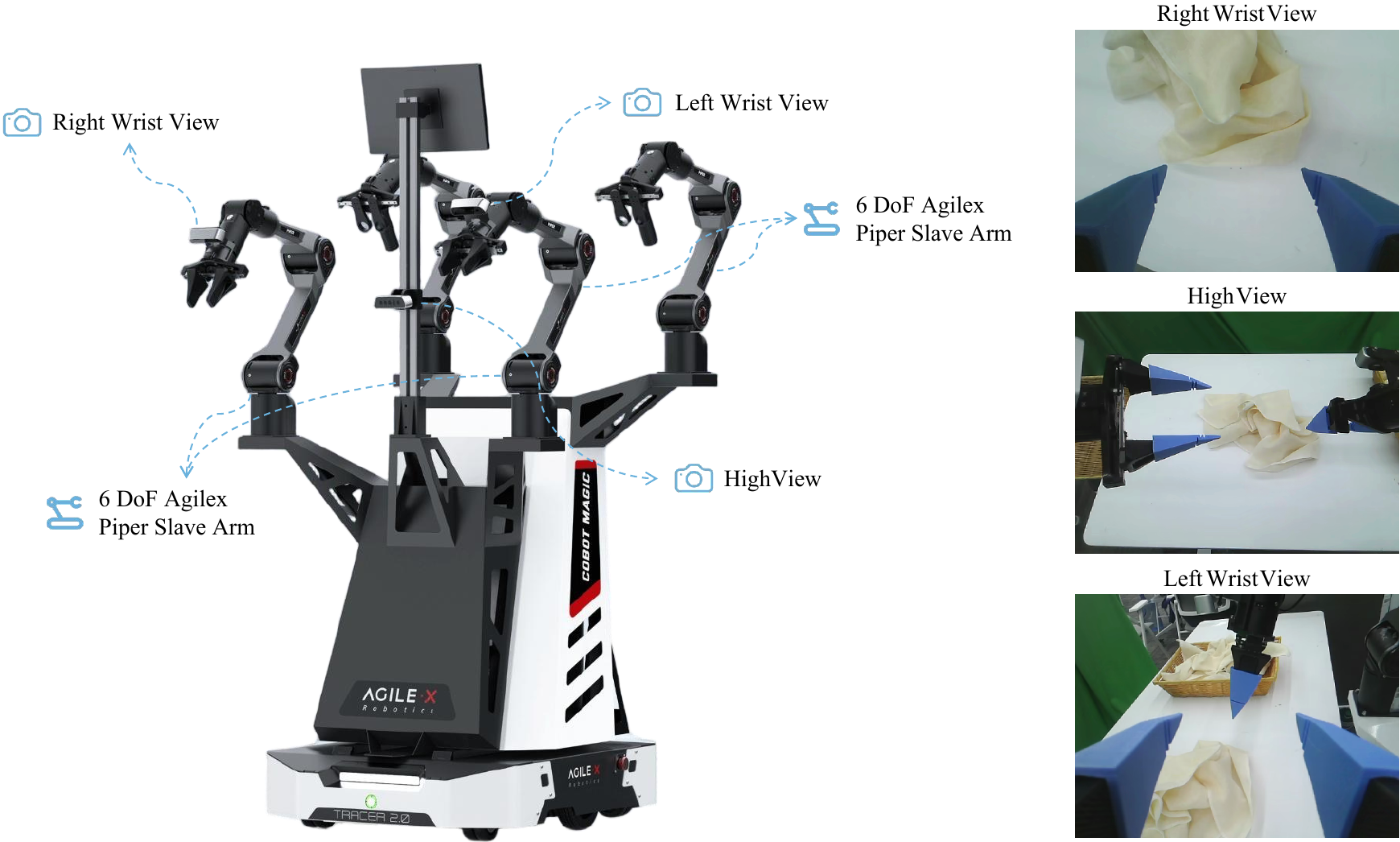}
    \caption{\textbf{Hardware setup for real-world experiments.} The system features a 6-DoF bimanual robot configuration controlled via an AgileX master-slave teleoperation interface. It is equipped with a global base camera and two wrist-mounted cameras to capture comprehensive visual observations alongside the 14-dimensional proprioceptive data.}
    \label{fig:hardware}
\end{figure}
\section{Real-World Implementation Details}
\label{sec:realworld_implementation}

\paragraph{Hardware Setup.}
The real-world data collection and policy deployment were conducted using an AgileX master-slave teleoperation system (illustrated in Fig.~\ref{fig:hardware}). The hardware platform utilizes a 6-Degree-of-Freedom (6-DoF) bimanual robot configuration.

\paragraph{Observation and Action Space.} 
To provide rich multimodal representations for both the ARM and downstream policies, the system integrates three distinct RGB camera perspectives: a \textit{High View} to capture the global context of the workspace, alongside \textit{Left} and \textit{Right Wrist Views} for egocentric, contact-rich visual feedback. Furthermore, both the proprioceptive state and the action space are 14-dimensional, comprising the continuous joint positions and gripper states of the bimanual manipulators.

\section{Folding Precision Evaluation Protocol}
We define a quantitative \textbf{folding precision score} ranging from 0 to 5 to evaluate the quality of towel folding results:



\begin{itemize}
    \item \textbf{5 points}: The folding task is fully completed, with a folding precision within 1 cm.
    \item \textbf{4 points}: The folding task is fully completed, with a folding precision between 1 cm and 2 cm.
    \item \textbf{3 points}: The folding task is fully completed, with a folding precision between 2 cm and 3 cm.
    \item \textbf{2 points}: The towel is successfully flattened, but the final folding is not completed, though partial folding steps are finished.
    \item \textbf{1 point}: The towel is successfully flattened, but no valid folding steps are performed.
    \item \textbf{0 points}: No task steps are successfully completed.
\end{itemize}